\documentclass[letterpaper]{article} % DO NOT CHANGE THIS
\usepackage{aaai24}  % DO NOT CHANGE THIS
\usepackage{times}  % DO NOT CHANGE THIS
\usepackage{helvet}  % DO NOT CHANGE THIS
\usepackage{courier}  % DO NOT CHANGE THIS
\usepackage[hyphens]{url}  % DO NOT CHANGE THIS
\usepackage{graphicx} % DO NOT CHANGE THIS
\usepackage{natbib}  % DO NOT CHANGE THIS AND DO NOT ADD ANY OPTIONS TO IT
\usepackage{caption} % DO NOT CHANGE THIS AND DO NOT ADD ANY OPTIONS TO IT
\usepackage{algorithm}
\usepackage{algorithmic}

\usepackage{newfloat}
\usepackage{listings}
\DeclareCaptionStyle{ruled}{labelfont=normalfont,labelsep=colon,strut=off} % DO NOT CHANGE THIS
\floatstyle{ruled}
\newfloat{listing}{tb}{lst}{}
\floatname{listing}{Listing}
\title{A Chain-of-Thought Prompting Approach with LLMs for Evaluating Students’ Formative Assessment Responses in Science}
\author{
    Clayton Cohn\textsuperscript{\rm 1},
    Nicole Hutchins\textsuperscript{\rm 1},
    Tuan Le\textsuperscript{\rm 2},
    Gautam Biswas\textsuperscript{\rm 1}
}
\affiliations{
    \textsuperscript{\rm 1}Vanderbilt University\\
    \textsuperscript{\rm 2}DePauw University\\
    \{clayton.a.cohn, nicole.m.hutchins, gautam.biswas\}@vanderbilt.edu, tuanle\_2025@depauw.edu
}

\begin{document}

\maketitle

\begin{abstract}
This paper explores the use of large language models (LLMs) to score and explain short-answer assessments in K-12 science. While existing methods can score more structured math and computer science assessments, they often do not provide explanations for the scores. Our study focuses on employing GPT-4 for automated assessment in middle school Earth Science, combining few-shot and active learning with chain-of-thought reasoning. Using a human-in-the-loop approach, we successfully score and provide meaningful explanations for formative assessment responses. A systematic analysis of our method's pros and cons sheds light on the potential for human-in-the-loop techniques to enhance automated grading for open-ended science assessments.
\end{abstract}

\section{Introduction} \label{sec:introduction}
Improvements in Science, Technology, Engineering, and Mathematics (STEM) education have accelerated the shift from teaching and assessing facts to developing students’ conceptual understanding and problem-solving skills \cite{ngss2013}. To foster students’ developing scientific ideas and reasoning skills, it is crucial to have assessments that reveal and support 
their progress \cite{harrisstemassessments}. Formative assessments play an important role in this endeavor, providing timely feedback and guidance when students face difficulties, which helps them to develop self-evaluation skills \cite{bloom1971formative}. However, the process of grading and generating personalized feedback from frequent formative assessments is time-consuming for teachers and susceptible to errors \cite{RODRIGUES201430, haudek2011harnessing}. %Computational approaches are needed. 

Large Language Models (LLMs) provide opportunities for automating short answer scoring \cite{funayama2023reducing} and providing feedback to help students overcome their difficulties \cite{morris2023using}. These approaches can also aid teachers in identifying students' difficulties and generating actionable information to support student learning. To our knowledge, there is very little research that combines automated formative assessment grading and feedback generation for science domains where understanding, reasoning, and explaining are key to gaining a deep understanding of scientific phenomena \cite{mao2018validation}. 

This paper develops an approach for human-in-the-loop LLM prompt engineering using in-context learning and chain-of-thought reasoning with GPT-4 to support automated analysis and feedback generation for formative assessments in a middle school Earth Science curriculum. We present our approach, discuss our results, evaluate the limitations of our work, and then propose future research in this area of critical need in K-12 STEM instruction.

\section{Background} \label{sec:background}
To understand the difficulties students face when learning science, teachers need to actively track students' developing knowledge \cite{wiley2020}. This is particularly important for open-ended, technology-enhanced learning environments that support students in their knowledge construction and problem-solving processes \cite{hutchins2023LAK}. In these environments, knowledge and skill development happen through system interactions that are difficult to monitor and interpret \cite{walkoe2017}. Formative assessments, evaluation, and feedback mechanisms aligned with target learning goals \cite{bloom1971formative}, can play a dual role: (1) help students recognize constructs that are important to learning, and (2) provide teachers with a deeper understanding of student knowledge and reasoning to better support their developing STEM ideas \cite{CIZEK20231}. However, grading formative assessments, particularly in K-12 STEM contexts, where students' responses may not be well-structured and may vary considerably in vocabulary and stylistic expression, is time-consuming and can result in erroneous scoring and incomplete feedback \cite{liu2016crater}. Moreover, grading these assessments at frequent intervals may become a burden rather than an aid to 
teachers. Very little research has examined effective mechanisms for generating automated grading and useful formative feedback for K-12 students that are aligned with classroom learning goals.

% \textbf{Formative Assessments in STEM, Specifically, Open-Ended, Problem-Based Learning:} 
% -Need to actively evaluate students’ developing science knowledge of concepts and construction processes to more deeply understand how they learn and difficulties they have\\
% -Formative assessments are often used in education to monitor student learning with the goal of giving them personalized feedback to improve their learning and for teachers to adapt their instruction to classroom needs\\
% - Could be particularly beneficial for problem-based / open-ended learning environments where students' construct their own knowledge as they provide an opportunity to monitor knowledge acquisition and potential misunderstandings, which are difficult to do in such curricular approaches
% -FAs are time consuming to grade, prone to bias / error in grading because students’ answers may not be properly structured, and they have a varied vocabulary and style of expression. This makes it difficult to develop rubrics for grading answers\\
% - Facilitation involves scoring but also, for formative evaluation, need feedback to improve learning process \\

Advances in natural language processing (NLP) have produced improved automated assessment scoring approaches to support teaching and learning (e.g., \citealp{adairAIEDassessments,WILSON2021104208}). Proposed methodologies include data augmentation \citep{cochran2023bimproving}, next sentence prediction \citep{wu2023matching}, prototypical neural networks \citep{zeng2023generalizable}, cross-prompt fine-tuning \citep{funayama2023reducing}, human-in-the-loop scoring via sampling responses % to be scored by humans intelligently
\citep{singla2022using}, and reinforcement learning \citep{liu2022giving}. While these methods have enjoyed varying degrees of success, a majority of these applications have targeted more structured mathematics and computer science tasks (i.e., tasks that can be solved formulaically), but their grading is different from scoring free-form short-answer responses by middle school students in science domains. %These methods also often stop at providing actionable insight into their scoring decisions. 
Data impoverishment concerns are common to educational data sets and a key consideration in applying these approaches to science assessments \citep{cochran2023improving}. The data needed for training our models is small, imbalanced, and non-canonical in terms of syntax and semantics, all of which may impact model performance \citep{cohn2020bert}. 

This research tackles several critical issues, namely: (1) grading open-ended, short-answer questions focused on science conceptual knowledge and reasoning, (2) utilizing LLMs to generate explanations aligned with specified learning objectives for both students and teachers and (3) addressing concerns related to data impoverishment. %In the next section, we provide our curricular context and study design, and detail our technical methods and analysis plan. 
%Our approach leverages active learning with chain of thought prompting that has previously been applied to logic word problems in algebra  \citealp{wei2022chain}. 
We hypothesize that our approach %, described above and detailed in the next section, 
supports automated scoring and explanation that (1) aligns with learning objectives and standards, (2) provides actionable insight to students, especially in addressing their difficulties, and (3) engages teachers in the scoring and explanation generation process to resolve discrepancies and support the learning goals. % (demonstrated successful in formative assessment technology development via co-design, c.f., \citealp{penuel2007formativecodesign}).

% \textbf{Our Approach}
% -OELE, problem-based learning in STEM, in particular in science domains that focus on scientific reasoning, e.g., drawing inferences and providing causal representations\\
% -difficulty tracking, assessing, and supporting developing STEM knowledge in these environments 
% -Formative assessments can be an important methods for learning in SPICE context (discuss multiple representations pursuant to this context)\\
% -formative assessments and the use of short-answer, reasoning questions limited due to difficulty in evaluating and generating actionable feedback
% -Scoring short answer/open-ended questions using LLMs has shown promise\\
% \indent -We can break down responses into science/reasoning subscores
% \indent -We can help understand why a model scores instances the way it does\\
% \indent -We can elicit explanations with our scores to help teachers provide actionable advice to help both teachers and students\\
% -human-in-the-loop methods also provide unique opportunity to integrate teacher insight and better align feedback with the learning objectives of the classroom
% -what is it that we want to accomplish with our short answering grading (need list)\\
% -

\section{Methods} \label{sec:methods}
This section presents our curricular context, study design, dataset, LLM, and the details of our approach. Additional information regarding the formative assessment questions, rubrics, prompts, and method application can be found in the GitHub repository\footnote{https://github.com/oele-isis-vanderbilt/EAAI24} along with test code and sample data. 

\subsection{Curricular context}
This paper evaluates formative assessments conducted in the context of Science Projects Integrating Computing and Engineering (SPICE), an NGSS-aligned middle school earth sciences water runoff curriculum. Spanning three weeks, the curriculum tasks students with redesigning their schoolyard to enhance functionalities, using surface materials that minimize water runoff post-storm within specified cost and accessibility constraints \citep{chiu2019}. We focus on formative assessments that are primarily linked to the %curriculum consists of three core units: (1) physical experiments of water runoff and 
conceptual understanding of water runoff and % to understand the primary concepts and their relations based on 
the conservation of matter principle \cite{hutchins2021isls}. %\textbf{} %; (2) computational modeling of the water runoff phenomenon; and (3) engineering design, in which students use their computational models to redesign their schoolyard to meet the specifications discussed above. While learning through this multiple, linked representations approach 
 % additional work is needed to support all students as they reason and transition through each linked domain \cite{Zhang2022book}. To do so, we aim to leverage existing assessments implemented throughout the XYZ curriculum as an opportunity to evaluate and provide feedback and guidance on students' developing science ideas and reasoning. 

\subsection{Study Design and Dataset} \label{subsec:study_design_dataset}
This study utilized assessment data from two Vanderbilt University-approved SPICE studies involving 270 students at a Southeastern U.S. public middle school. Data was removed for non-consenting participants and some data was missing because of absences and incomplete submissions. We used evidence-centered design (ECD) \cite{mislevy2006} to align the assessments with the learning objectives of the SPICE curriculum. %integrated science, computing, and engineering
%Our research targets the grading of short-answer science questions that allow for clear applications of conceptual knowledge and ask students to explain their scientific reasoning. 

\begin{figure}[tb]
    \centering
    \includegraphics[width=1\linewidth]{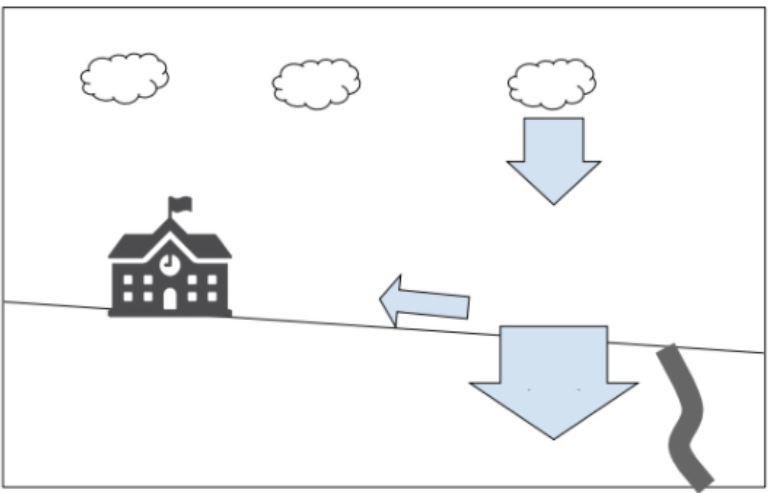}
    \caption{The fictitious student's conceptual model used by students to answers the assessment questions.}
    \label{fig:assessment}
\end{figure}

For this paper, we selected three questions that required students to analyze a pictorial model of water runoff (illustrated in Figure \ref{fig:assessment}) and apply their conceptual knowledge and scientific reasoning to evaluate and explain the correct and incorrect components of the model. %that involved student evaluations of an example conceptual model created by a fictitious student:
Each question was scored for at least one conceptual knowledge item, i.e., a correct application of a scientific fact. For example, in Q3, students had to identify that the arrow size representing total absorption was incorrect. Q2 and Q3 also required scoring students' scientific reasoning, i.e., the use of scientific principles to explain an answer. For Q3, students could invoke the conservation principle to explain that the absorption arrow could not be larger than the rainfall arrow. %, because of the conservation principle that rainfall should be equal to the absorption and runoff in the conceptual model. 
The rubric assigned 1 point (conceptual) for Q1. Q2 and Q3 were scored for 4 points (2 items, 1 conceptual and 1 reasoning point for each item). For Q3, there were exactly 2 errors in the model. For Q2, students could choose from more than two correct phenomena, which resulted in differences in the grading results that we discuss later. %constraining the open-ended solution construction space. 

\subsection{Model} \label{sec:Models}
Ever since OpenAI released ChatGPT\footnote{https://openai.com/blog/chatgpt} (a chatbot driven by the foundation model GPT-3.5) in November 2022, %. Since ChatGPT's release, 
LLMs have received a tremendous amount of attention. Their ability to %emulate human writing while
compose paper outlines, expository essays, and screenplays, has made the use of ChatGPT ubiquitous across academia, business, and news media. In March 2023, OpenAI released GPT-4\footnote{https://openai.com/research/gpt-4} \citep{openai2023gpt4}, which is largely considered the current state-of-the-art for LLMs \citep{openai2023gpt4,%touvron2023llama,
zhao2023survey}. For this reason, we chose to use GPT-4 as the LLM to develop and evaluate our approach.

\subsection{Approach}
Brown et al. (\citeyear{brown2020language}) demonstrated that LLMs could ``learn" from a few labeled instances in the prompt via \textit{in-context learning} (ICL). Unlike fine-tuning, which requires expensive parameter updates and may result in decreased performance for previously known tasks \citep{mosbach2020stability}, ICL uses the labeled instances in the prompt to generate text during inference that bypasses traditional training. This means that by simply changing the prompts, the same language model can be used across domains, tasks, and datasets without the need to modify the network's parameters. Wei et al. (\citeyear{wei2022chain}) extended this work by providing \textit{chain-of-thought} (CoT) reasoning in the labeled instances. In contrast to a traditional ICL instance that only offers a question and its corresponding answer, CoT provides a reasoning chain with the answer. This helps the model generate correct inferences, and this reasoning is included in the model's response along with the answer.

Eliciting reasoning is particularly useful for formative assessment scoring in science, where the open-ended nature of the questions can make scoring alignment difficult even between humans. Rather than generating a score only, CoT prompting elicits an explanation for the LLM's response, enabling teachers to offer informed feedback to students. Alternatively, teachers can refine the rubric to improve grading for subsequent assessments. The model's reasoning can also be used to identify specific causes of misalignment between the model and the teacher, which can then be leveraged to improve model output. 

\textit{Active learning} \citep{tan2023does,ren2021survey} %,cohn1994improving} 
takes a human-in-the-loop approach to improving model training, where the human as an ``oracle" % (i.e., the human) 
is consulted to label additional instances for inclusion in the next training iteration. By integrating CoT reasoning and active learning, educators or researchers can scrutinize instances with incorrect predictions to identify recurring patterns leading to the model's errors across multiple instances. These patterns can be reintroduced into the prompt using CoT reasoning to rectify discrepancies between the model's assessment and the human scorer.
 % Additionally, this process is informative to rubric development and refinement, as it can help locate ambiguities in the rubric and formative assessment questions the model encountered that may similarly create confusion among students. 
Moreover, combining CoT with active learning assists teachers and researchers in rectifying human errors in the initial scoring. This is particularly relevant when the humans confirm that the model's scoring predictions are accurate.

\begin{figure*}[tb]
    \centering
    \includegraphics[width=0.75\textwidth]{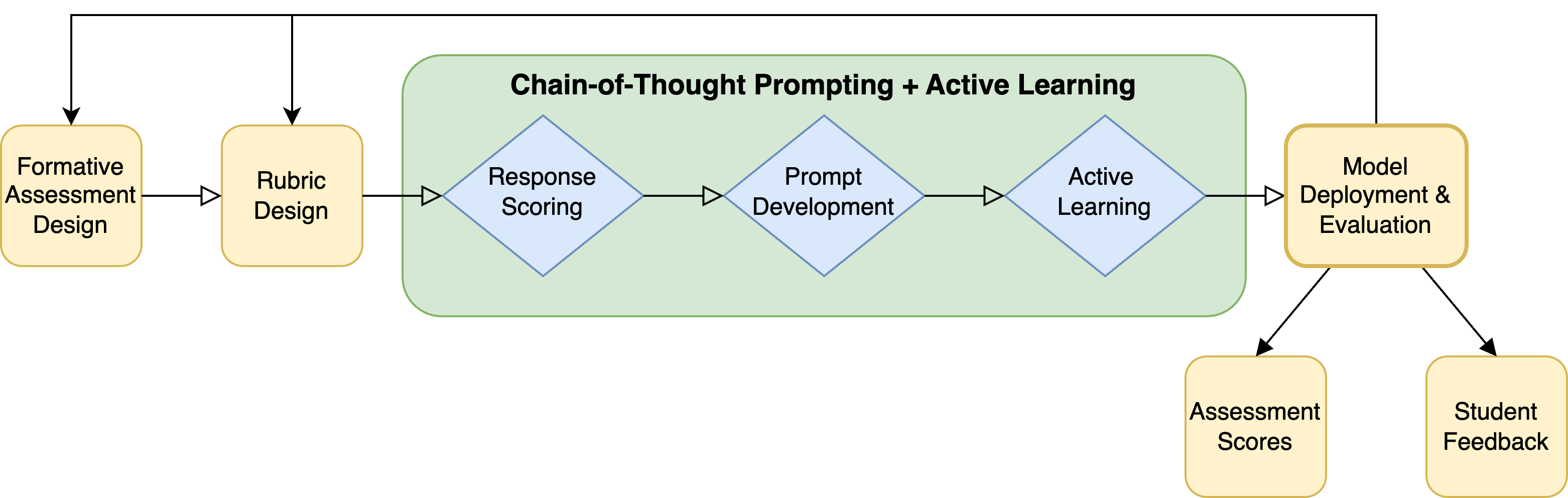}
    \caption{Our Chain-of-Thought Prompting + Active Learning approach. The green box encapsulates this process, where each of the blue diamonds is a step in that process. Yellow boxes represent the process's application to the classroom.}
    \label{fig:approach}
\end{figure*}

%In this paper, we % use a human-in-the-loop approach to score and explain students' short answer  formative assessment responses in the Earth Science domain. We 
We employ the inter-rater reliability (IRR) process to pinpoint scoring disagreements that may challenge the model, addressing them through CoT prompting. Active learning is then utilized to identify recurrent issues in the model's alignment with the human scorers, and instances embodying these patterns are incorporated into the prompt with reasoning chains to correct the alignment. Once active learning concludes, the model is deployed for scoring new formative assessment responses through inference, accompanied by CoT reasoning to generate student feedback, and when needed, refining rubrics and formative assessment questions. Figure \ref{fig:approach} provides a comprehensive overview of our approach. % and was performed for each of the three sets of formative assessment responses (i.e., each of the three questions) we evaluated in this paper.

\subsubsection{Response Scoring.} 
Two of this paper's authors independently scored a randomly chosen 20\% of the student responses for each of the three formative assessment questions using the rubric.
% after which IRR procedures were conducted. 
Next, while conducting IRR, instances where the humans both agreed and disagreed on students' scores were collected and % instances are 
included in the initial prompt. Particular attention was paid to the misalignments between the graders that caused multiple instances to be scored differently before consensus was reached. To achieve consensus, the two reviewers discussed each scoring disagreement until they reached a consensus on how that particular instance should be scored. % during IRRthe reasons why the graders disagreed before consensus, especially . 
The agreed-upon instances acted as ``ground truth'' exemplars for the model to initially align itself with the human scorers. The instances where there were disagreements were used to pinpoint specific reasons for misalignment between the human scorers during IRR. We expected that the model might encounter the same misalignments during its scoring. This process was repeated for each of the three questions until Cohen's $k > 0.7$ was achieved across all subscores for each question, after which one of this paper's authors scored the full set of student responses. For this work, all students' responses were manually graded to ensure accuracy while evaluating our method. Disagreements were resolved manually by the humans to form a consensus (described above). This consensus was used to align the LLM responses via CoT reasoning. In future work, as we collect more data, we will use the LLM to automatically score students' responses and evaluate samples of the LLM's generations to ensure accuracy. Furthermore,  we refrained from updating the rubric during Active Learning; however, we intend to investigate this aspect in future research.

Before developing the initial prompt, we partitioned the dataset into training (80\%) and testing (20\%) instances %(80\% and 20\%, respectively) 
for the three sets of formative assessment responses. The training set played a dual role in prompt development. Initially, few-shot examples were selected to construct the prompt, while the instances not utilized for few-shot learning served as a validation set for refining the prompt during active learning. Due to token limitations and the time cost for instance labeling, only a limited number of labeled instances were included in the prompt. As discussed in later sections, an excessive number of instances in the prompt can lead to overfitting. Furthermore, it is important that the validation set during active learning is sufficiently large to ensure accurate identification of scoring trends. In this paper, the validation-to-training set ratio was $\approx$43:1.

\subsubsection{Prompt Development.} For prompting, we %first instructed the model on its task. We 
opted for the \textit{persona pattern} \citep{white2023prompt}, where the model was instructed to adopt the persona of a middle school teacher evaluating students' formative assessment question responses. The prompt also provided the model with the formative assessment question and rubric, and the model was instructed to use the rubric to score students' responses. % pursuant to that rubric. 
The rubric also provided the model with the format to output its responses to improve readability and allow for programmatic parsing of the model's generations. 

Next, we incorporated ground truth examples into the prompt, complemented by CoT reasoning clarifying the reasons for awarding or not awarding points for each subscore. Following this, a comparable CoT input was included for instances where human scorers diverged in their assessments. This aimed at aligning the model with the IRR consensus, particularly when instances posed challenges similar to those faced by human reviewers in achieving consensus. %, using CoT reasoning to explain why or why not the student should be awarded a point if that sticking point is encountered at inference time. 
For all labeled instances in the prompt, we used the following CoT reasoning template: % to tie the model's scoring to the rubric: 
evidence in the student's response + reference to the rubric + score. We used quotations from the student’s response as evidence, tying it back to the rubric, and providing a score and explanation to the model; e.g., ``The student says X. The rubric states Y. Based on the rubric, the student earned a score of Z."  This approach mirrored the original CoT publication \citep{wei2022chain}, where algebraic word problems were broken down step-by-step to help the model arrive at the correct solution.

Additional labeled instances were added to the prompt as needed to balance the individual subscores. %and approximate an equal distribution. %While we did not use data augmentation in this work, prior work demonstrated augmenting underrepresented classes to balance the dataset can improve model performance considerably \citep{cochran2022improving,cochran2023improving,cochran2023bimproving}. In our case, we simply selected additional instances from the training set and added them to the list of labeled instances in the prompt. 
However, this was constrained by the small and imbalanced nature of our dataset. %Because we were balancing across 4 subscores per student response (for Q2 and Q3), it was not always possible to achieve an exact balance.
While investigating the effect that data balance has on the LLM's performance is outside the scope of this work, in previous work (using a subset of the dataset used in this paper), we demonstrated that data balancing often improved language model performance \cite{cochran2022improving}. For Q2 and Q3, balancing across 4 subscores was difficult, as adding one more instance to augment one subscore inherently affected the balance across the other subscores. %Often, the exact set of subscores needed to 
Sometimes, achieving a perfect balance was not possible in the training set, but we included at least one positive and one negative instance across all subscores for each question's prompt.

\subsubsection{Active Learning.} Validation set instances %(i.e., the portion of the training set not used in the prompt) 
were fed through the model with the initial prompt and few-shot examples, and a researcher performed error analysis to discern patterns in the incorrect LLM generations. Specifically, we noted the reason for each incorrect scoring prediction and the faulty reasoning chains that caused the model to mislabel several instances. % in the model's explanation. Each 
These reasoning chains were chosen %as a candidate for a new few-shot instance 
as additional examples to add to the prompt, and CoT was used to correct the model's reasoning errors. %This allowed for potentially correcting several wrongly scored instances with each additional labeled example and its corresponding CoT reasoning. 
Candidate instances were prioritized for prompt inclusion based on the degree to which their reasoning errors caused other inaccurate model predictions, which resulted in correcting several wrongly scored instances.

There were only a few incorrectly predicted scores in the validation set for Q1, so all of those instances were added to the prompt during Active Learning. For Q2 and Q3, the researcher identified the $n$ most useful instances, where $n$ was defined as the minimum number of instances in the validation set that simultaneously addressed all of the LLM's reasoning errors and maintained data balance. This caused some overfitting, so we will experiment with 1-shot active learning to help mitigate this in future work. For all instances added to the prompt during Active Learning, we used CoT to correct the model's faulty reasoning chains. We also rebalanced the few-shot instances across subscores during Active Learning to maintain data balance. In previous work, we showed that balancing training data to create a uniform label distribution can improve performance \cite{cochran2022improving}. Other works have suggested balancing to achieve the true distribution of the dataset's labels \cite{min2022rethinking}.

In general, active learning can be performed until one of several stopping conditions is triggered: (1) the model achieves convergence, i.e., it no longer produces any incorrect validation scores; (2) the model predicts more validation scores incorrectly than in previous iterations, i.e., it overfits; and (3) there are not enough instances remaining in the validation set to achieve acceptable data balance in the prompt.

%\begin{itemize}
 %   \item The model no longer produces any incorrect validation generation scores (i.e., convergence)
%    \item The model predicts more validation scores incorrectly than in previous iterations (i.e., overfitting)
 %   \item There are not enough instances remaining in the validation dataset to achieve acceptable data balance in the prompt
%    \item Other real-world constraints are realized (e.g., API call limits, API token limits, context length limits, financial preclusions due to cost, etc.)
%\end{itemize}

To test our method, we performed one iteration of active learning for each of the three formative assessment questions. For each subscore of a formative assessment question, we first identified scoring error trends, i.e., are model scoring errors mainly caused by false negatives (underscoring) or false positives (overscoring)? This alerted us to the ``direction" in which we needed to guide the model to better align with the human scorers. We then examined the content of the incorrect validation set generations to identify common causes of incorrect scoring. We chose the most frequently occurring model reasoning error (i.e., the error that caused the model to wrongly predict the greatest number of validation set instances), and picked one of these instances to insert back into the prompt.

For example, with the \textit{Runoff Arrow Direction} subscore in Q3, we found that the ratio of the model's false positive to false negative predictions was 5:2. Additionally, we found that the cause of more than half of the false positives was due to the model awarding students a point for mentioning that the arrows in the diagram needed to change direction. This was incorrect because only the runoff arrow needed to change direction. %, as the other arrows in the diagram were already pointing in the correct direction. 
To correct the model's reasoning error, we chose one of the incorrect validation instances that included this reasoning error, inserted it into the prompt, and used CoT reasoning to help correct the model's reasoning error going forward.

%\subsubsection{Deployment and Evaluation.}
%Once satisfied with the model's performance on the validation set, the final prompt is ready for \textit{Deployment and Evaluation}. During deployment, the prompt can be used to score students' formative assessment responses, and the model's CoT reasoning can be used by humans to provide feedback to students as to why each student was or was not awarded points for a particular formative assessment question. Additionally, humans can also use the information learned in the \textit{Response Scoring}, \textit{Prompt Development}, and \textit{Active Learning} processes to refine the formative assessment questions and rubric. As mentioned, the two authors of this paper who conducted IRR realized that Q2 and its rubric need to be rewritten to be more clear, as the wording created confusion for students, researchers, and the LLM. 
\section{Results} \label{sec:Results}

We evaluated our method by comparing our model performance to the held-out test set across 4 implementations: three incremental baselines, and our Chain-of-Thought Prompting + Active Learning approach. We started with a \textit{Zero-Shot} baseline, where the rubric is included in the prompt, but no labeled examples were present. We then used a \textit{Few-Shot} baseline, where we provided the model with labeled instances in the prompt, but the labeled instances only consisted of numerical scores (i.e., no CoT reasoning). Our third and final baseline, \textit{Few-Shot, CoT}, added CoT reasoning to the few-shot instances. Last, we employed our Chain-of-Thought Prompting + Active Learning approach and compared it to the three baselines. Evaluating our approach across these incremental baselines allowed us to examine the effects of adding specific parts of the pipeline and to understand the degree to which each component contributed to the model's ability to score and explain formative assessment responses. %We present our results in the Results section. 

To compare implementations, we chose the Macro F1-Score and Cohen's Quadratic Weighted Kappa (QWK) \citep{cohen1968weighted} metrics. The F1-Score is prevalent in the literature for evaluating overall model performance. Macro F1 was chosen, specifically, due to our dataset's imbalance across subscores. Often, scientific reasoning subscores are heavily weighted towards the negative class (i.e., a large majority of the students do not demonstrate scientific reasoning). Cohen's QWK was chosen because it is widely used in the automated essay scoring (AES) literature \citep{singh2023h,singla2022using}. %,riordan2017investigating}. 
Unlike traditional Cohen's $k$ \citep{cohen1960coefficient}, Cohen's QWK accounts for the degree of disagreement, making it well-suited for ordinal data. We included accuracy for reference, but we do not use it in our actual performance comparisons.

%In the Discussion, we dive deeper into the model's scoring explanations to examine how helpful they are in providing actionable insight with respect to students’ science knowledge and reasoning abilities, and we identify the areas where the model needs improvement. To this end, we conduct a thematic analysis on the model's incorrect scoring predictions and highlight several common reasons for the model's inability to accurately score and explain the students' responses. Additionally, we compare the model's incorrect scores while using our approach to scores labeled by the second human scorer (i.e., the scorer who did not do the final scoring during IRR), and we report the degree to which the second human scorer agreed with the model versus the first human scorer. 

Model performance comparisons for each of the three formative assessment questions are shown in Tables~\ref{tab:q1_results},~\ref{tab:q2_results}, and~\ref{tab:q3_results}. \\ \\ %For each question, we compare our Chain-of-Thought Prompting + Active Learning approach to the \textit{Zero-Shot}, \textit{Few-Shot}, and \textit{Few-Shot, CoT} baselines discussed in the Methods section.
\noindent
\textbf{Question 1:}
Q1 asked students what the different-sized arrows in the diagram meant. % and had one possible point for science concepts via the ``Arrow Size" subscore. 
A student received a point for correctly identifying that the diagram used the size of the arrows to represent the quantity of water (concept: ``Arrow Size'', see Table \ref{tab:q1_results}). %Results for Q1 are displayed in Table \ref{tab:q1_results}.

\begin{table}[tb]
    \centering
    \begin{tabular}{|l|l|l|l|l|}
         \hline
         \textbf{Q1 Arrow Size}& n & Acc & F1 & QWK\\
         \hline
         Zero-Shot & 0 & 0.87 & 0.84 & 0.68 \\
         Few-Shot & 4 & 1.00 & \textbf{1.00} & \textbf{1.00} \\
         Few-Shot, CoT & 4 & 0.96 & 0.95 & 0.89 \\
         CoT + AL & 12 & 0.98 & 0.97 & 0.95 \\
         \hline
    \end{tabular}
    \caption{Performance comparisons for the Q1 \textit{Arrow Size} subscore. For all questions, the best-performing scoring implementation is in bold for each metric, for each subscore (and total score). $n$ refers to the number of few-shot instances used in the prompt.}
    \label{tab:q1_results}
\end{table}

Q1 took 2 rounds of IRR for the human scorers to reach a consensus. %and it was the simplest of the three questions, as there was only 
The grading involved scoring for one possible point and no science reasoning subscores. GPT-4 aligned with the human scorer to a ``moderate" degree (QWK $>=$ 0.6) \citep{mchugh2012interrater} even in a zero-shot setting. Once labeled instances were added, the model achieved a perfect score on the test set. When CoT reasoning was provided, performance decreased for both Macro F1 and QWK, as the reasoning chains initially caused the model to deviate from the human scorer. Once active learning was performed, however, much of that performance gap was closed due to the additional few-shot instances and model reasoning error corrections. \\ \\
%\vspace{1.2em}
\noindent 
\textbf{Question 2:} Q2 asked students to identify two things that the diagram did well for 4 possible points: 2 for science concepts, and 2 for science reasoning. For the science concepts subscores, the student received a point for \textit{Arrow Direction} if he or she correctly identified that the diagram did a good job of showing that water originated from the sky in the form of rain, some water was absorbed, or some resulted in runoff. For \textit{Arrow Size}, students received a point if they discussed that the diagram did a good job of using arrow size to represent the water amount. Each of the \textit{Arrow Direction} and \textit{Arrow Size} subscores also included an additional possible point if students demonstrated scientific reasoning in their responses (see Table \ref{tab:q2_results}). %Performance results for Q2 are shown in Table \ref{tab:q2_results} for each subscore and the total score.

% Overall results
% Analysis doc
\begin{table}[tb]
    \centering
    \begin{tabular}{|l|l|l|l|l|}
         \hline
         \textbf{Q2 Arrow Direction}& n & Acc & F1 & QWK\\
         \hline
         Zero-Shot & 0 & 0.91 & 0.89 & 0.78 \\
         Few-Shot & 5 & 0.87 & 0.79 & 0.60\\
         Few-Shot, CoT & 5 & 0.98 & \textbf{0.98} & \textbf{0.95}\\
         CoT + AL & 10 & 0.98 & \textbf{0.98} & \textbf{0.95}\\
         \hline
         \hline
         \textbf{Q2 Arr. Dir., Reasoning}& n & Acc & F1 & QWK\\
         \hline
         Zero-Shot & 0 & 0.92 & \textbf{0.73} & \textbf{0.47}\\
         Few-Shot & 5 & 0.89 & 0.67 & 0.36\\
         Few-Shot, CoT & 5 & 0.91 & 0.70 & 0.41\\
         CoT + AL & 10 & 0.92 & 0.65 & 0.3\\
         \hline
         \hline
         \textbf{Q2 Arrow Size}& n & Acc & F1 & QWK\\
         \hline
         Zero-Shot & 0 & 0.77 & 0.69 & 0.39\\
         Few-Shot & 5 & 0.77 & 0.69 & 0.39\\
         Few-Shot, CoT & 5 & 0.91 & 0.88 & 0.77\\
         CoT + AL & 10 & 0.94 & \textbf{0.92} & \textbf{0.83}\\
         \hline
         \hline
         \textbf{Q2 Arr. Sz., Reasoning}& n & Acc & F1 & QWK\\
         \hline
         Zero-Shot & 0 & 0.96 & 0.82 & 0.65\\
         Few-Shot & 5 & 0.98 & \textbf{0.90} & \textbf{0.79}\\
         Few-Shot, CoT & 5 & 0.94 & 0.77 & 0.55\\
         CoT + AL & 10 & 0.96 & 0.82 & 0.65\\
         \hline
         \hline
         \textbf{Q2 Total Score}& n & Acc & F1 & QWK\\
         \hline
         Zero-Shot & 0 & 0.60 & 0.59 & 0.65\\
         Few-Shot & 5 & 0.53 & 0.52 & 0.55\\
         Few-Shot, CoT & 5 & 0.75 & \textbf{0.80} & 0.80\\
         CoT + AL & 10 & 0.85 & 0.79 & \textbf{0.87}\\
         \hline
    \end{tabular}
    \caption{Performance comparisons for Question 2.}
    \label{tab:q2_results}
\end{table}

Q2 science concepts subscores (i.e., \textit{Arrow Direction} and \textit{Arrow Size}) saw their best performance (or tied for best performance) using the full Chain-of-Thought Prompting + Active Learning method. The science reasoning subscores' performances decreased as additional components of the method were added, i.e., after CoT and active learning were introduced. Overall, the total score was best when the complete method was used, as this resulted in the highest QWK value.

Q2 was the most difficult for the human scorers to agree on and it required three separate IRR rounds to achieve consensus. Some of the difficulty in scoring may be attributed to the open-ended nature of the question. There are multiple ideas in the conceptual model that are correct, but students were only asked to identify two things the diagram explained well. Many students responded vaguely, and several students provided both correct and incorrect statements in the same response. These types of ambiguous instances were difficult for the human scorers to agree on even when they awarded points based on the rubric. It seems the LLM encountered the same types of issues.

%The model struggled to score the student responses for several of the same reasons the researchers found it difficult to reach consensus during IRR, even after CoT reasoning was deployed to correct the misalignments. 
Consider a student whose Q2 response was, ``[the arrow represents] the amount of absorption". Arguably, the student understood that the model's arrows corresponded to the quantity of water. However, the absorption arrow in the diagram was incorrect (it was larger than the rainfall arrow, so it violated the law of conservation of matter). Because the question asked for examples of things the diagram does a \textit{good} job of, and the absorption arrow was incorrect, both reviewers felt responses like this one should not receive a point for \textit{Arrow Size} even though the student may understand that arrow size corresponds to the amount of water. During our active learning validation run, the model incorrectly awarded several points to these types of responses. When we attempted to use CoT to correct the model's reasoning error, the model began to mislabel other instances it had previously scored correctly, i.e., there was overfitting. Ultimately, the researchers agreed that both the Q2 question wording and the Q2 rubric need to be rewritten to provide clearer guidance to the students. \\ \\
\noindent 
%\vspace{1.2em}
\textbf{Question 3:} Q3 asked students to list two erroneous things they would change in the conceptual model diagram. Like Q2, 4 total points were assigned to Q3: 2 for science concepts and 2 for scientific reasoning. The science concepts subscores were: (1) \textit{Runoff Direction}, where the student received a point if he or she indicated that the runoff arrow was pointing the wrong direction (uphill) %and needed to change direction (downhill); 
and (2) \textit{Arrow Size}, where a point was awarded if the student mentioned that the arrow sizes needed to change and adhere to the law of conservation of matter. Similar to Q2, students got additional points if they demonstrated correct scientific reasoning in their responses (see Table \ref{tab:q3_results}).

\begin{table}[tb]
    \centering
    \begin{tabular}{|l|l|l|l|l|}
         \hline
         \textbf{Q3 Runoff Direction}& n & Acc & F1 & QWK\\
         \hline
         Zero-Shot & 0 & 0.89 & 0.88 & 0.77 \\
         Few-Shot & 5 & 0.91 & 0.90 & 0.80 \\
         Few-Shot, CoT & 5 & 0.92 & \textbf{0.92} & \textbf{0.84} \\
         CoT + AL & 9 & 0.89 & 0.88 & 0.75\\
         \hline
         \hline
         \textbf{Q3 Run. Dir., Reasoning}& n & Acc & F1 & QWK\\
         \hline
         Zero-Shot & 0 & 0.94 & 0.89 & 0.79\\
         Few-Shot & 5 & 0.94 & 0.91 & 0.82\\
         Few-Shot, CoT & 5 & 0.94 & 0.92 & 0.83 \\
         CoT + AL & 9 & 0.98 & \textbf{0.97} & \textbf{0.94}\\
         \hline
         \hline
         \textbf{Q3 Arrow Size}& n & Acc & F1 & QWK\\
         \hline
         Zero-Shot & 0 & 0.87 & 0.83 & 0.67\\
         Few-Shot & 5 & 0.89 & 0.87 & 0.73\\
         Few-Shot, CoT & 5 & 0.85 & 0.83 & 0.65\\
         CoT + AL & 9 & 0.92 & \textbf{0.92} & \textbf{0.83}\\
         \hline
         \hline
         \textbf{Q3 Arr. Sz., Reasoning}& n & Acc & F1 & QWK\\
         \hline
         Zero-Shot & 0 & 0.98 & 0.90 & 0.79\\
         Few-Shot & 5 & 1.00 & \textbf{1.00} & \textbf{1.00}\\
         Few-Shot, CoT & 5 & 0.94 & 0.82 & 0.64\\
         CoT + AL & 9 & 1.00 & \textbf{1.00} & \textbf{1.00}\\
         \hline
         \hline
         \textbf{Q3 Total Score}& n & Acc & F1 & QWK\\
         \hline
         Zero-Shot & 0 & 0.74 & \textbf{0.80} & 0.85 \\
         Few-Shot & 5 & 0.75 & 0.73 & 0.87 \\
         Few-Shot, CoT & 5 & 0.75 & 0.71 & 0.79\\
         CoT + AL & 9 & 0.81 & \textbf{0.80} & \textbf{0.90} \\
         \hline
    \end{tabular}
    \caption{Performance comparisons for Question 3.}
    \label{tab:q3_results}
\end{table}

All Q3 subscores (science concepts and scientific reasoning) improved performance across both metrics (except Macro F1 for total score) after we added the few-shot examples. When CoT was added, performance increased for both \textit{Runoff Direction} subscores but decreased substantially for both \textit{Arrow Size} subscores. %, where the \textit{Arrow Size Reasoning} subscore, in particular, saw a considerable drop. 
We saw similar behavior in the Q1 \textit{Arrow Size} subscore, where adding CoT reasoning caused the model to become misaligned with the human. Once the Active Learning component was added, however, all subscores except \textit{Runoff Direction} achieved their best performance across both metrics. \textit{Runoff Direction} achieved its best performance when CoT was added, but was overfit during active learning. Unlike Q2, where the best-performing subscores were the science concepts, both science reasoning subscores did better than their science concepts counterparts for both metrics.

For Q3, the human scorers achieved consensus quickly after 1 round of IRR, and only one issue caused multiple scoring disagreements. The model's reasoning errors for the scientific reasoning subscores were easily addressed via CoT (relative to Q2). A major model reasoning error for Q3 was that it tended to cite the same piece of evidence to justify awarding points for different subscores (i.e., it overscored). This was a disagreement with the human scorers, but we did not include it in the initial prompt or few-shot CoT reasoning chains. Once this model reasoning error was addressed during active learning, the issue was largely mitigated, and performance improved across the board. \\ \\
 %\vspace{1.2em}
 \noindent
\textbf{Summary:} Across all questions, the model's scoring mostly aligned with the human scorers. Of the 11 subscores and total scores, 9 of them saw ``strong" agreement or better (QWK $>=$ 0.8) at some point in the process (i.e., across the 4 implementations: 3 baselines and our Chain-of-Thought Prompting + Active Learning approach). 4 subscores achieved ``almost perfect" (QWK $>$ 0.9) agreement. All subscores except one (Q2 Arrow Direction Reasoning) saw a Macro F1 of 0.90 or greater at some point in the process. Importantly, we also demonstrated that both CoT reasoning and active learning run the risk of overfitting, particularly when applied to the less complex science concepts questions (e.g., Q1 \textit{Arrow Size} and Q3 \textit{Runoff Direction}) and the more ambiguous scientific reasoning questions (e.g., Q2 \textit{Arrow Direction Reasoning} and Q2 \textit{Arrow Size Reasoning}). It should also be noted that the level of agreement during IRR may provide a ballpark expectation of model performance, as we found questions that were easier for the human scorers to agree on were also easier for the model to correctly align with the human scorers. Similarly, in questions where the human scorers had difficulty achieving consensus, the model had difficulty with scoring. %, many times struggling with the same types of questions the human scorers grappled with during IRR. However, 
More research needs to be done to evaluate this quantitatively.  

% \section{Discussion} \label{sec:discussion}

\section{Comparing Model and Human Performance} 
We applied inductive coding \citep{charmaz2006} to evaluate performance and identify future directions to improve our human-in-the-loop approach. First, the lead author (not involved in rubric creation and scoring) reviewed all instances in which the model and the human coder disagreed and identified agreement with the model in 3 out of the 22 disagreements (1 conceptual disagreement, 2 reasoning disagreements). The research team reviewed the results to evaluate what may have caused scoring errors and to identify potential future directions for improvement. During the review process, the team created memos of key findings \citep{hatch2002}. The team compared the memos and came up with three key themes for improvement in future work: 
 
\begin{enumerate}
    \item \textbf{Need for Additional Mechanisms to Target Model Deficiencies}: Differences in scoring identified that the model showed a tendency to overfit in some cases. For instance, if the CoT got too granular, the model demonstrated issues that were related to keywords such as ``because" (e.g., the model identified it as a demonstration of reasoning), ``arrow size” (e.g., the model assumed that use of the terminology indicated a correct application even if correct attributions were not made to the scientific process), and vocabulary definitions (e.g., the model did not realize ``run off” and ``runoff” were identical). In a small set of cases, the model cited a student's faulty logic to justify awarding a point for a response and reused the same piece of evidence to award points for both concept identification and reasoning; %, or the same evidence to justify both reasoning scores;
    \item \textbf{Ability to Leverage the Model to Support Rubric Refinements}: Comparing reviewer and model differences for Q2 helped identify limitations in the original rubric for such an open-ended question. Utilizing the results and the explanations provided by the model, this human-in-the-loop approach can benefit teachers and researchers in refining the rubrics and scoring mechanisms to better support instruction and student learning; and
    \item \textbf{Resolve Unexplained Model Applications}: In some cases, %in which the model cited positive predictions, 
    the model did not follow CoT reasoning and did not provide evidence of its positive predictions even though all positive prompt instances provided this evidence. This may be a potential limitation in the approach to providing feedback for positive performances. 
\end{enumerate}

%In nearly all instances, the model selects good evidence (i.e., the same evidence a human would consider when scoring), and the model is able to tie that evidence to the rubric and score accordingly. 
Overall, our approach was successful, but the instances discussed above provide opportunities for future work to improve model output, rubric development, and sometimes even reworking questions to make them clearer. 

\section{Conclusion and Future Implications} \label{sec:conclusion}
In this paper, we employed a Chain-of-Thought Prompting + Active Learning approach for scoring and explaining formative assessment question responses in a middle school Earth Science curriculum. %We compared our method's scoring efficacy across three incremental baselines using two metrics (Macro F1 and Cohen's QWK). 
Our results show that GPT-4, CoT reasoning, and active learning can be effectively leveraged toward accurate grading of science formative assessments. In several cases, the model achieved ``almost perfect" alignment with humans. The model generated relevant evidence linked to the rubric to help explain its scoring, which could benefit students and teachers. %Our method also highlights potential formative assessment question and rubric ambiguities that may cause confusion for both the students and the LLM. 
We also analyzed the model's weaknesses and identified several areas for improving LLM-based assessments.

\noindent
\textbf{Limitations:} With LLM approaches, ethical concerns arise with regard to privacy, bias, and hallucinations \citep{zhuo2023red}, and these concerns are amplified when they are deployed in high-stakes environments (e.g., classrooms with children). %GPT-4, specifically, raises additional questions due to its opacity and private ownership. Not all teachers and researchers have access to GPT-4 due to its exclusivity, token limits, and API access and cost\footnote{For reference, we spent a total of \$91.62 using the OpenAI API for testing, refining, and evaluating our method.}. Unfortunately, there is a considerable gap between open and closed LLMs in terms of performance \citep{touvron2023llama}, and model size plays a large role in customizing LLMs \citep{kaplan2020scaling}. Initially, we used a considerably smaller, open-source LLM Falcon-7B-Instruct\footnote{https://huggingface.co/tiiuae/falcon-7b-instruct}, which did not produce adequate model generations to support our approach, and it was difficult to improve performance with CoT prompting. In future work, we will study smaller, customized models that are effective for grading formative assessments.
In addition, while CoT has been shown to improve model performance over traditional ICL, the degree to which the reasoning chains guide the model's decision-making (if at all) is still an open question \cite{turpin2023language}. %Trying to elicit LLM responses that ``align" with humans can often lead to a decrease in overall model performance and have other unintended consequences \citep{wolf2023fundamental}. This is referred to as the \textit{alignment problem}\footnote{https://openai.com/blog/our-approach-to-alignment-research} and is currently an active field of research. An additional drawback to CoT prompting (and ICL as a whole) is that the task description, rubric, and few-shot CoT examples must accompany each instance during inference, which can drive up API costs. 
Our results also show that CoT and active learning can lead to overfitting, in particular, with simpler, easier-to-define subproblems. In these cases, LLM approaches may be overkill, as \citet{moore2023assessing} recently demonstrated. Rule-based methods outperformed GPT-4 in detecting common item-writing flaws in student-generated multiple-choice questions. %Other limitations to our approach are discussed in the Results and Discussion sections.

\noindent 
\textbf{Looking to the Future:}
% Unlike fine-tuning, our method does not require traditional training or network updates. This makes our approach generalizable, as a single model can be applied to multiple tasks, domains, and datasets by swapping out the prompts. In future work, we will investigate more complex science formative assessment questions such as \textit{causal reasoning} \citep{hughes2019automatic} %and \textit{``what-if" analyses} \citep{}, and work 
% to extend our instructional support capabilities. % via agent development and refinement. %We will also explore other prompt refinements such as instructing the prompt to not rely on keywords like ``because" and to not reuse evidence across subscores. 
% We will explore refinement to the active learning process to reduce overfitting, including benefits leveraged via data augmentation \cite{cochran2022improving}. %to generate a single instance that balances subscores and corrects one model reasoning error per subscore in a single shot. 
% This will allow us to perform active learning one instance at a time and potentially reduce the model's tendency to overfit. 
Anecdotally, in an interview with middle school science teachers who implemented the curriculum, the teachers identified the potential benefits of these explanations as tools to inform students on where to go next in their learning, as opposed to assigning performance scores. We aim to extend this partnership with classroom teachers to mold the LLM's output to best fit their needs, and investigate how we can best use our method to evaluate students' learning performance and improve students' learning. As we continue to refine our approach, we hope these enhancements will pave the way for more effective and efficient LLM applications in science education.

\section*{Acknowledgments}
% NSF AI Institute Grant No. DRL-2112635\\
% Cyberlearning grant IIS-2017000

This work is supported by the National Science Foundation under awards DRL-2112635 and IIS-2017000. Any opinions, findings, conclusions, and recommendations in this paper are those of the authors and do not necessarily reflect the views of the National Science Foundation.

\bibliography{main}

\begin{thebibliography}{44}
\providecommand{\natexlab}[1]{#1}

\bibitem[{Adair et~al.(2023)Adair, Pedro, Gobert, and Segan}]{adairAIEDassessments}
Adair, A.; Pedro, M.~S.; Gobert, J.; and Segan, E. 2023.
\newblock Real-Time AI-Driven Assessment and Scaffolding that Improves Students' Mathematical Modeling during Science Investigations.
\newblock In Wang, N.; Rebolledo-Mendez, G.; Matsuda, N.; Santos, O.~C.; and Dimitrova, V., eds., \emph{Artificial Intelligence in Education}, 202--216. Cham: Springer Nature Switzerland.
\newblock ISBN 978-3-031-36272-9.

\bibitem[{Bloom, Madaus, and Hastings(1971)}]{bloom1971formative}
Bloom, B.; Madaus, G.; and Hastings, J. 1971.
\newblock \emph{Handbook on Formative and Summative Evaluation of Student Learning}.
\newblock New York: McGraw-Hill.

\bibitem[{{Brown} et~al.(2020){Brown}, {Mann}, {Ryder}, {Subbiah}, {Kaplan}, {Dhariwal}, {Neelakantan}, {Shyam}, {Sastry}, {Askell}, {Agarwal}, {Herbert-Voss}, {Krueger}, {Henighan}, {Child}, {Ramesh}, {Ziegler}, {Wu}, {Winter}, {Hesse}, {Chen}, {Sigler}, {Litwin}, {Gray}, {Chess}, {Clark}, {Berner}, {McCandlish}, {Radford}, {Sutskever}, and {Amodei}}]{brown2020language}
{Brown}, T.~B.; {Mann}, B.; {Ryder}, N.; {Subbiah}, M.; {Kaplan}, J.; {Dhariwal}, P.; {Neelakantan}, A.; {Shyam}, P.; {Sastry}, G.; {Askell}, A.; {Agarwal}, S.; {Herbert-Voss}, A.; {Krueger}, G.; {Henighan}, T.; {Child}, R.; {Ramesh}, A.; {Ziegler}, D.~M.; {Wu}, J.; {Winter}, C.; {Hesse}, C.; {Chen}, M.; {Sigler}, E.; {Litwin}, M.; {Gray}, S.; {Chess}, B.; {Clark}, J.; {Berner}, C.; {McCandlish}, S.; {Radford}, A.; {Sutskever}, I.; and {Amodei}, D. 2020.
\newblock {Language Models are Few-Shot Learners}.
\newblock \emph{arXiv e-prints}, arXiv:2005.14165.

\bibitem[{Charmaz(2006)}]{charmaz2006}
Charmaz, K. 2006.
\newblock \emph{Constructing grounded theory: A practical guide through qualitative analysis}.
\newblock Sage.

\bibitem[{Chiu et~al.(2019)Chiu, McElhaney, Zhang, Biswas, Fried, Basu, Alozie, and Hong}]{chiu2019}
Chiu, J.; McElhaney, K.; Zhang, N.; Biswas, G.; Fried, R.; Basu, S.; Alozie, N.; and Hong, J. 2019.
\newblock A Principled Approach to NGSS-aligned Curriculum Development Integrating Science, Engineering, and Computation: A Pilot Study.
\newblock In \emph{NARST Annual International Conference}. NARST.

\bibitem[{Cizek and Lim(2023)}]{CIZEK20231}
Cizek, G.~J.; and Lim, S.~N. 2023.
\newblock Formative assessment: an overview of history, theory and application.
\newblock In Tierney, R.~J.; Rizvi, F.; and Ercikan, K., eds., \emph{International Encyclopedia of Education (Fourth Edition)}, 1--9. Oxford: Elsevier, fourth edition edition.
\newblock ISBN 978-0-12-818629-9.

\bibitem[{Cochran, Cohn, and Hastings(2023)}]{cochran2023bimproving}
Cochran, K.; Cohn, C.; and Hastings, P. 2023.
\newblock Improving NLP model performance on small educational data sets using self-augmentation.
\newblock In \emph{Proceedings of the 15th International Conference on Computer Supported Education (2023, to appear)}.

\bibitem[{Cochran et~al.(2022)Cochran, Cohn, Hutchins, Biswas, and Hastings}]{cochran2022improving}
Cochran, K.; Cohn, C.; Hutchins, N.; Biswas, G.; and Hastings, P. 2022.
\newblock Improving automated evaluation of formative assessments with text data augmentation.
\newblock In \emph{International Conference on Artificial Intelligence in Education}, 390--401. Springer.

\bibitem[{Cochran et~al.(2023)Cochran, Cohn, Rouet, and Hastings}]{cochran2023improving}
Cochran, K.; Cohn, C.; Rouet, J.~F.; and Hastings, P. 2023.
\newblock Improving Automated Evaluation of Student Text Responses Using GPT-3.5 for Text Data Augmentation.
\newblock In \emph{International Conference on Artificial Intelligence in Education}, 217--228. Springer.

\bibitem[{Cohen(1960)}]{cohen1960coefficient}
Cohen, J. 1960.
\newblock A coefficient of agreement for nominal scales.
\newblock \emph{Educational and psychological measurement}, 20(1): 37--46.

\bibitem[{Cohen(1968)}]{cohen1968weighted}
Cohen, J. 1968.
\newblock Weighted kappa: nominal scale agreement provision for scaled disagreement or partial credit.
\newblock \emph{Psychological bulletin}, 70(4): 213.

\bibitem[{Cohn(2020)}]{cohn2020bert}
Cohn, C. 2020.
\newblock \emph{BERT efficacy on scientific and medical datasets: a systematic literature review}.
\newblock DePaul University.

\bibitem[{Funayama et~al.(2023)Funayama, Asazuma, Matsubayashi, Mizumoto, and Inui}]{funayama2023reducing}
Funayama, H.; Asazuma, Y.; Matsubayashi, Y.; Mizumoto, T.; and Inui, K. 2023.
\newblock Reducing the Cost: Cross-Prompt Pre-finetuning for Short Answer Scoring.
\newblock In \emph{International Conference on Artificial Intelligence in Education}, 78--89. Springer.

\bibitem[{Harris et~al.(2023)Harris, Wiebe, Grover, and Pellegrino}]{harrisstemassessments}
Harris, C.; Wiebe, E.; Grover, S.; and Pellegrino, J. 2023.
\newblock \emph{Classroom-based STEM assessment: Contemporary issues and perspectives}.
\newblock Community for Advancing Discovery Research in Education (CADRE). Education Development Center, Inc.

\bibitem[{Hatch(2002)}]{hatch2002}
Hatch, J.~A. 2002.
\newblock \emph{Doing qualitative research in education settings}.
\newblock SUNY Press.

\bibitem[{Haudek et~al.(2011)Haudek, Kaplan, Knight, Long, Merrill, Munn, Nehm, Smith, and Urban-Lurain}]{haudek2011harnessing}
Haudek, K.~C.; Kaplan, J.~J.; Knight, J.; Long, T.; Merrill, J.; Munn, A.; Nehm, R.; Smith, M.; and Urban-Lurain, M. 2011.
\newblock Harnessing technology to improve formative assessment of student conceptions in STEM: forging a national network.
\newblock \emph{CBE—Life Sciences Education}, 10(2): 149--155.

\bibitem[{Hutchins and Biswas(2023)}]{hutchins2023LAK}
Hutchins, N.; and Biswas, G. 2023.
\newblock Using Teacher Dashboards to Customize Lesson Plans for a Problem-Based, Middle School STEM Curriculum.
\newblock In \emph{LAK23: 13th International Learning Analytics and Knowledge Conference}, LAK2023, 324–332. New York, NY, USA: Association for Computing Machinery.
\newblock ISBN 9781450398657.

\bibitem[{Hutchins et~al.(2021)Hutchins, Basu, McElhaney, Chiu, Fick, Zhang, and Biswas}]{hutchins2021isls}
Hutchins, N.~M.; Basu, S.; McElhaney, K.; Chiu, J.; Fick, S.; Zhang, N.; and Biswas, G. 2021.
\newblock Coherence across conceptual and computational representations of students’ scientific models.
\newblock In \emph{The International Society of the Learning Sciences Annual Meeting 2021}. International Society of the Learning Sciences (ISLS).

\bibitem[{Liu et~al.(2022)Liu, Stephan, Nie, Piech, Brunskill, and Finn}]{liu2022giving}
Liu, E.; Stephan, M.; Nie, A.; Piech, C.; Brunskill, E.; and Finn, C. 2022.
\newblock Giving Feedback on Interactive Student Programs with Meta-Exploration.
\newblock \emph{Advances in Neural Information Processing Systems}, 35: 36282--36294.

\bibitem[{Liu et~al.(2016)Liu, Rios, Heilman, Gerard, and Linn}]{liu2016crater}
Liu, O.~L.; Rios, J.~A.; Heilman, M.; Gerard, L.; and Linn, M.~C. 2016.
\newblock Validation of automated scoring of science assessments.
\newblock \emph{Journal of Research in Science Teaching}, 53(2): 215--233.

\bibitem[{Mao et~al.(2018)Mao, Liu, Roohr, Belur, Mulholland, Lee, and Pallant}]{mao2018validation}
Mao, L.; Liu, O.~L.; Roohr, K.; Belur, V.; Mulholland, M.; Lee, H.-S.; and Pallant, A. 2018.
\newblock Validation of automated scoring for a formative assessment that employs scientific argumentation.
\newblock \emph{Educational Assessment}, 23(2): 121--138.

\bibitem[{McHugh(2012)}]{mchugh2012interrater}
McHugh, M.~L. 2012.
\newblock Interrater reliability: the kappa statistic.
\newblock \emph{Biochemia medica}, 22(3): 276--282.

\bibitem[{Min et~al.(2022)Min, Lyu, Holtzman, Artetxe, Lewis, Hajishirzi, and Zettlemoyer}]{min2022rethinking}
Min, S.; Lyu, X.; Holtzman, A.; Artetxe, M.; Lewis, M.; Hajishirzi, H.; and Zettlemoyer, L. 2022.
\newblock Rethinking the role of demonstrations: What makes in-context learning work?
\newblock \emph{arXiv preprint arXiv:2202.12837}.

\bibitem[{Mislevy and Haertel(2006)}]{mislevy2006}
Mislevy, R.~J.; and Haertel, G.~D. 2006.
\newblock Implications of Evidence-Centered Design for Educational Testing.
\newblock \emph{Educational Measurement: Issues and Practice}, 25(4): 6--20.

\bibitem[{Moore et~al.(2023)Moore, Nguyen, Chen, and Stamper}]{moore2023assessing}
Moore, S.; Nguyen, H.~A.; Chen, T.; and Stamper, J. 2023.
\newblock Assessing the Quality of Multiple-Choice Questions Using GPT-4 and Rule-Based Methods.
\newblock In \emph{European Conference on Technology Enhanced Learning}, 229--245. Springer.

\bibitem[{Morris et~al.(2023)Morris, Crossley, Holmes, Ou, McNamara, and Dascalu}]{morris2023using}
Morris, W.; Crossley, S.; Holmes, L.; Ou, C.; McNamara, D.; and Dascalu, M. 2023.
\newblock Using Large Language Models to Provide Formative Feedback in Intelligent Textbooks.
\newblock In \emph{International Conference on Artificial Intelligence in Education}, 484--489. Springer.

\bibitem[{Mosbach, Andriushchenko, and Klakow(2020)}]{mosbach2020stability}
Mosbach, M.; Andriushchenko, M.; and Klakow, D. 2020.
\newblock On the stability of fine-tuning bert: Misconceptions, explanations, and strong baselines.
\newblock \emph{arXiv preprint arXiv:2006.04884}.

\bibitem[{NGSS(2013)}]{ngss2013}
NGSS. 2013.
\newblock \emph{Next Generation Science Standards: For States, By States}.
\newblock The National Academies Press.

\bibitem[{{OpenAI}(2023)}]{openai2023gpt4}
{OpenAI}. 2023.
\newblock {GPT-4 Technical Report}.
\newblock \emph{arXiv e-prints}, arXiv:2303.08774.

\bibitem[{Ren et~al.(2021)Ren, Xiao, Chang, Huang, Li, Gupta, Chen, and Wang}]{ren2021survey}
Ren, P.; Xiao, Y.; Chang, X.; Huang, P.-Y.; Li, Z.; Gupta, B.~B.; Chen, X.; and Wang, X. 2021.
\newblock A survey of deep active learning.
\newblock \emph{ACM computing surveys (CSUR)}, 54(9): 1--40.

\bibitem[{Rodrigues and Oliveira(2014)}]{RODRIGUES201430}
Rodrigues, F.; and Oliveira, P. 2014.
\newblock A system for formative assessment and monitoring of students' progress.
\newblock \emph{Computers \& Education}, 76: 30--41.

\bibitem[{Singh et~al.(2023)Singh, Pupneja, Mital, Shah, Bawkar, Gupta, Kumar, Kumar, Gupta, and Shah}]{singh2023h}
Singh, S.; Pupneja, A.; Mital, S.; Shah, C.; Bawkar, M.; Gupta, L.~P.; Kumar, A.; Kumar, Y.; Gupta, R.; and Shah, R.~R. 2023.
\newblock H-AES: Towards Automated Essay Scoring for Hindi.
\newblock \emph{arXiv preprint arXiv:2302.14635}.

\bibitem[{Singla et~al.(2022)Singla, Krishna, Shah, and Chen}]{singla2022using}
Singla, Y.~K.; Krishna, S.; Shah, R.~R.; and Chen, C. 2022.
\newblock Using sampling to estimate and improve performance of automated scoring systems with guarantees.
\newblock In \emph{Proceedings of the AAAI Conference on Artificial Intelligence}, volume 36 (11), 12835--12843.

\bibitem[{Tan et~al.(2023)Tan, Lin, Lang, Chen, Ga{\v{s}}evi{\'c}, Du, and Buntine}]{tan2023does}
Tan, W.; Lin, J.; Lang, D.; Chen, G.; Ga{\v{s}}evi{\'c}, D.; Du, L.; and Buntine, W. 2023.
\newblock Does informativeness matter? Active learning for educational dialogue act classification.
\newblock In \emph{International Conference on Artificial Intelligence in Education}, 176--188. Springer.

\bibitem[{Turpin et~al.(2023)Turpin, Michael, Perez, and Bowman}]{turpin2023language}
Turpin, M.; Michael, J.; Perez, E.; and Bowman, S.~R. 2023.
\newblock Language Models Don't Always Say What They Think: Unfaithful Explanations in Chain-of-Thought Prompting.
\newblock \emph{arXiv preprint arXiv:2305.04388}.

\bibitem[{Walkoe, Wilkerson, and Elby(2017)}]{walkoe2017}
Walkoe, J.; Wilkerson, M.; and Elby, A. 2017.
\newblock Technology-Mediated Teacher Noticing: A Goal for Classroom Practice, Tool Design, and Professional Development.
\newblock In \emph{Proceedings of the 12th International Conference on Computer Supported Collaborative Learning (CSCL) 2017}. International Society of the Learning Sciences.

\bibitem[{{Wei} et~al.(2022){Wei}, {Wang}, {Schuurmans}, {Bosma}, {Ichter}, {Xia}, {Chi}, {Le}, and {Zhou}}]{wei2022chain}
{Wei}, J.; {Wang}, X.; {Schuurmans}, D.; {Bosma}, M.; {Ichter}, B.; {Xia}, F.; {Chi}, E.; {Le}, Q.; and {Zhou}, D. 2022.
\newblock {Chain-of-Thought Prompting Elicits Reasoning in Large Language Models}.
\newblock \emph{arXiv e-prints}, arXiv:2201.11903.

\bibitem[{White et~al.(2023)White, Fu, Hays, Sandborn, Olea, Gilbert, Elnashar, Spencer-Smith, and Schmidt}]{white2023prompt}
White, J.; Fu, Q.; Hays, S.; Sandborn, M.; Olea, C.; Gilbert, H.; Elnashar, A.; Spencer-Smith, J.; and Schmidt, D.~C. 2023.
\newblock A prompt pattern catalog to enhance prompt engineering with chatgpt.
\newblock \emph{arXiv preprint arXiv:2302.11382}.

\bibitem[{Wiley et~al.(2020)Wiley, Dimitriadis, Bradford, and Linn}]{wiley2020}
Wiley, K.~J.; Dimitriadis, Y.; Bradford, A.; and Linn, M.~C. 2020.
\newblock From Theory to Action: Developing and Evaluating Learning Analytics for Learning Design.
\newblock In \emph{Proceedings of the Tenth International Conference on Learning Analytics \& Knowledge}, LAK '20, 569–578. New York, NY, USA: Association for Computing Machinery.
\newblock ISBN 9781450377126.

\bibitem[{Wilson et~al.(2021)Wilson, Ahrendt, Fudge, Raiche, Beard, and MacArthur}]{WILSON2021104208}
Wilson, J.; Ahrendt, C.; Fudge, E.~A.; Raiche, A.; Beard, G.; and MacArthur, C. 2021.
\newblock Elementary teachers’ perceptions of automated feedback and automated scoring: Transforming the teaching and learning of writing using automated writing evaluation.
\newblock \emph{Computers \& Education}, 168: 104208.

\bibitem[{Wu et~al.(2023)Wu, He, Liu, Liu, and Zhai}]{wu2023matching}
Wu, X.; He, X.; Liu, T.; Liu, N.; and Zhai, X. 2023.
\newblock Matching exemplar as next sentence prediction (mensp): Zero-shot prompt learning for automatic scoring in science education.
\newblock In \emph{International Conference on Artificial Intelligence in Education}, 401--413. Springer.

\bibitem[{Zeng et~al.(2023)Zeng, Li, Guan, Ga{\v{s}}evi{\'c}, and Chen}]{zeng2023generalizable}
Zeng, Z.; Li, L.; Guan, Q.; Ga{\v{s}}evi{\'c}, D.; and Chen, G. 2023.
\newblock Generalizable Automatic Short Answer Scoring via Prototypical Neural Network.
\newblock In \emph{International Conference on Artificial Intelligence in Education}, 438--449. Springer.

\bibitem[{Zhao et~al.(2023)Zhao, Zhou, Li, Tang, Wang, Hou, Min, Zhang, Zhang, Dong et~al.}]{zhao2023survey}
Zhao, W.~X.; Zhou, K.; Li, J.; Tang, T.; Wang, X.; Hou, Y.; Min, Y.; Zhang, B.; Zhang, J.; Dong, Z.; et~al. 2023.
\newblock A survey of large language models.
\newblock \emph{arXiv preprint arXiv:2303.18223}.

\bibitem[{Zhuo et~al.(2023)Zhuo, Huang, Chen, and Xing}]{zhuo2023red}
Zhuo, T.~Y.; Huang, Y.; Chen, C.; and Xing, Z. 2023.
\newblock Red teaming chatgpt via jailbreaking: Bias, robustness, reliability and toxicity.
\newblock \emph{arXiv preprint arXiv:2301.12867}, 12--2.

\end{thebibliography}

\end{document}